\pdfoutput=1

\documentclass[11pt]{article}

\usepackage[final]{acl}

\usepackage{times}
\usepackage{latexsym}

\usepackage[T1]{fontenc}

\usepackage[utf8]{inputenc}

\usepackage{microtype}

\usepackage{inconsolata}

\usepackage{graphicx}
\usepackage{amsmath}
\usepackage{amssymb}
\usepackage{bbm}
\usepackage{xcolor}
\usepackage{color}
\usepackage{float}
\usepackage{booktabs}
\usepackage{multirow}
\usepackage{makecell}

\usepackage{colortbl}
\usepackage{pifont}
\usepackage{graphicx}
\usepackage{subcaption}
\usepackage{algorithm}
\usepackage{algorithmicx}
\usepackage{algpseudocode}
\usepackage{fdsymbol}
\usepackage{pdfpages}

\newcommand{\nb}{\texttt{Nb}}
\newcommand{\bi}{\texttt{Bi}}
\newcommand{\up}{\texttt{Up}}
\newcommand{\dn}{\texttt{Dn}}
\newcommand{\internal}{\texttt{In}}
\newcommand{\ext}{\texttt{Ex}}
\newcommand{\m}{\texttt{M}}
\newcommand{\nm}{\texttt{Nm}}

\definecolor{orange}{HTML}{FF9933}
\definecolor{light-blue}{HTML}{3399FF}
\definecolor{dark-blue}{HTML}{3333FF}
\title{A Systematic Study of Compositional Syntactic Transformer Language Models}

\author{Yida Zhao$^{1,2}$, Hao Xve$^{1,2}$, Xiang Hu$^3$, Kewei Tu$^{1,2}$ \thanks{~Corresponding author}\\
  School of Information Science and Technology, ShanghaiTech University$^1$ \\
  Shanghai Engineering Research Center of Intelligent Vision and Imaging$^2$\\ 
  Ant Group$^3$\\
    {\tt \{zhaoyd2023,xvehao,tukw\}@shanghaitech.edu.cn, aaron.hx@antgroup.com}\\
 }

\begin{document}
\maketitle
\begin{abstract}
Syntactic language models (SLMs) enhance Transformers by incorporating syntactic biases through the modeling of linearized syntactic parse trees alongside surface sentences. This paper focuses on compositional SLMs that are based on constituency parse trees and contain explicit bottom-up composition of constituent representations. We identify key aspects of design choices in existing compositional SLMs and propose a unified framework encompassing both existing models and novel variants.
We conduct a comprehensive empirical evaluation of all the variants in our framework across language modeling, syntactic generalization, summarization, dialogue, and inference efficiency. Based on the experimental results, we make multiple recommendations on the design of compositional SLMs. Our code is released at {\url{https://github.com/zhaoyd1/compositional_SLMs}}.
\end{abstract}

\section{Introduction}

Transformer language models (LMs) have achieved remarkable success on various NLP tasks~\cite{devlin-etal-2019-bert, radford2019language, NEURIPS2020_1457c0d6, ouyang2022training}. While the Transformer architecture~\cite{vaswani2017attention} is highly powerful, it lacks the inductive bias of syntactic structures, which is believed to be critical for effective generalization~\cite{everaert2015structures}. Syntactic language models (SLMs)~\cite{qian-etal-2021-structural, yoshida-oseki-2022-composition, sartran-etal-2022-Transformer, murty-etal-2023-pushdown, zhao-etal-2024-dependency, hu-etal-2024-generative} incorporate such syntactic biases into Transformers with a straightforward method: modeling linearized syntactic parse trees along with the surface sentences.  

A major class of SLMs, which we call compositional SLMs, are based on constituency parse trees and contain explicit composition of sub-constituent representations to form constituent representations \cite{sartran-etal-2022-Transformer, yoshida-oseki-2022-composition, hu-etal-2024-generative}. 
These compositional SLMs differ in some key aspects, including the form of parse trees, the tree linearization strategy, the composition function, and the attention masking scheme. However, the specific impact of these aspects on SLM performance in language modeling and downstream tasks remains under-explored.

In this paper, we propose a unified framework of compositional SLMs that encompasses all these aspects. Our framework subsumes not only existing models as special cases but also more than ten novel variants.
We then conduct a systematic empirical comparison of all the variants in language modeling, syntactic generalization, summarization and dialogue (as two representative downstream tasks), and inference efficiency. The experimental results indicate that, compared with the Transformer LM baseline, compositional SLMs may underperform in language modeling, but the top-performing variants demonstrate significantly improved syntactic generalization, summarization, and dialogue performance, thus confirming the benefit of incorporating syntactic biases and explicit composition. We also observe significant performance and efficiency differences between the variants and make several recommendations on the design choices of compositional SLMs, such as discouraging sub-constituent masking and encouraging the combination of a specialized composition function and binary parse trees.

In summary, our contributions are two-fold:
\begin{itemize}
    \item We identify key aspects of design choices seen in existing compositional SLMs and propose a unified framework encompassing both existing models and novel variants.
    \item We conduct a comprehensive empirical evaluation of all the variants within our framework across a range of metrics, which leads to multiple recommendations on the design of compositional SLMs.
\end{itemize}

\section{Compositional SLM: A Framework}
This section defines a framework that subsumes existing compositional SLMs as special cases. We start with an overview of the framework before delving into details of four key aspects of design choices.

\begin{figure}[tb]
    \centering
    \includegraphics[width=\columnwidth]{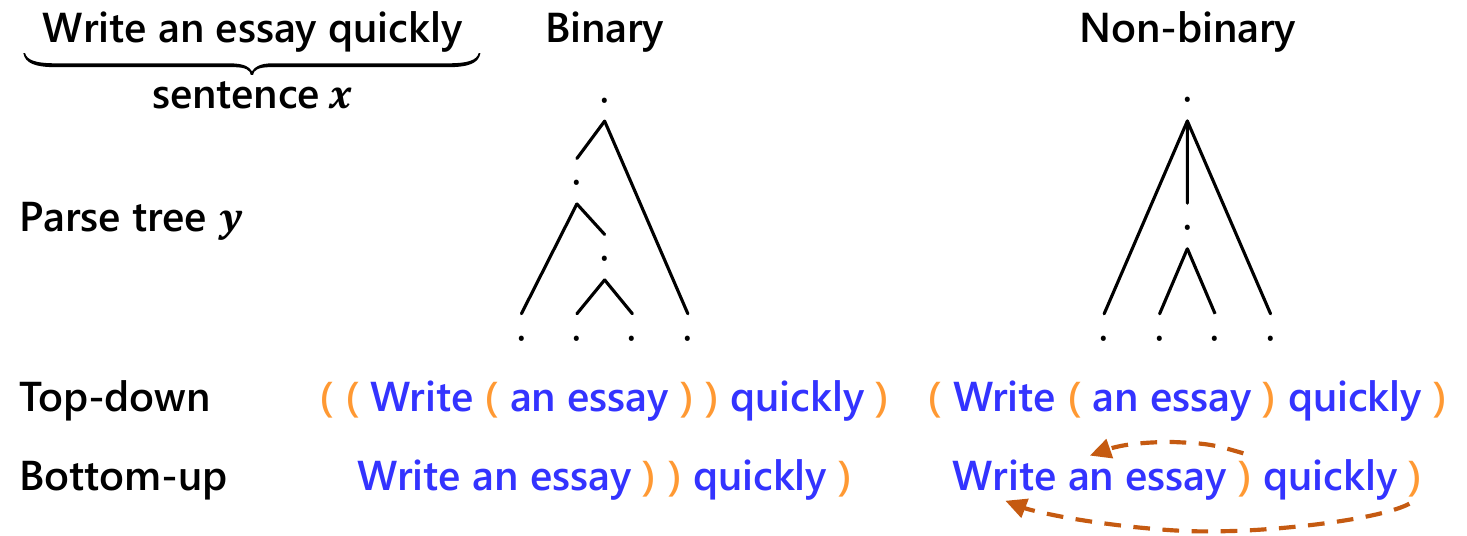}
    \caption{An example sentence, its binary and non-binary parse trees, and their linearizations produced by the two methods. For the bottom-up linearization of the non-binary tree, arcs are used to point from ")" actions to their corresponding start positions.}
    \label{fig:parse}
\end{figure}

A compositional syntactic language model (SLM) defines a joint distribution of sentences $\mathbf{x}$ and their constituency parse trees $\mathbf{y}$. For simplicity, we focus on unlabeled constituency trees in this paper. Following previous work on generative parsing and SLMs~\cite{dyer-etal-2016-recurrent, choe-charniak-2016-parsing, sartran-etal-2022-Transformer}, we define a sequence of actions $\boldsymbol{a} = (a_0, a_1, ..., a_{L-1})$ of length $L$ that construct $(\mathbf{x},\mathbf{y})$ in a left-to-right manner, where $a_i$ is an action that either generates a token in $\mathbf{x}$ or indicates bracketing within a parse tree in $\mathbf{y}$. We say $\boldsymbol{a}$ is a \emph{linearization} of $(\mathbf{x},\mathbf{y})$. Figure ~\ref{fig:parse} shows examples of two types of constituency parse trees and two linearization methods, which will be explained in section~\ref{sec:branching} and \ref{sec:top-down} respectively. The joint probability of $(\mathbf{x},\mathbf{y})$ can then be computed in an autoregressive way:
\[
    p(\mathbf{x}, \mathbf{y}) = p(\boldsymbol{a}) = \prod_i p(a_i | \boldsymbol{a}_{<i})
\]
A Transformer is utilized to model the autoregressive generation of action sequence $\boldsymbol{a}$. 

A compositional SLM also leverages explicit composition that calculates a composed representation for each nonterminal constituent in a constituency parse tree from the representations of its sub-constituents. 
How the composed representation is calculated and integrated into the Transformer will be discussed in section~\ref{sec:composition}.
Since the information of the sub-constituents is contained in the composed representation, once the composition is done, the sub-constituents can be optionally masked in the Transformer for subsequent action generation (section~\ref{sec:masks}).


\subsection{Parse Tree Binarization}\label{sec:branching}
A linguistically defined constituency tree $\mathbf{y}$ is generally a non-binary tree. However, previous studies on SLMs~\cite{murty-etal-2023-pushdown, hu-etal-2024-generative} often model binarized parse trees, highlighting the potential practical benefits of binarization. 
We consider both options in our framework.
(i) \textbf{Non-binary trees}, denoted as \textbf{\nb}. We eliminate all unary chains from any constituency trees, so structures like "\textit{( quickly )}" are simplified to "\textit{quickly}". (ii) \textbf{Binary trees}, denoted as \textbf{\bi}. We convert any non-binary tree into the Chomsky normal form using left binarization. 

\subsection{Linearization Methods} \label{sec:top-down}
We consider two linearization methods for converting a constituency parse tree into an action sequence: top-down and bottom-up. \textbf{Top-down} linearization~\cite{dyer-etal-2016-recurrent, sartran-etal-2022-Transformer, yoshida-oseki-2022-composition}, denoted as \textbf{\dn}, constructs a tree using pre-order traversal from the root to the terminals, with each nonterminal visited right before its children. In contrast, \textbf{bottom-up} linearization~\cite{hu-etal-2024-generative}, denoted as \textbf{\up}, constructs a tree using post-order traversal from the leaf terminals to the root, with each nonterminal visited right after all its children are visited. 

The choice between top-down and bottom-up linearization results in different action spaces. In top-down linearization, there are three types of actions: (i) opening a nonterminal (indicating the start of a new constituent), represented by "(", (ii) generating a terminal (a new token), represented directly by the token, and (iii) closing a nonterminal (indicating the end of the current constituent), represented by ")". On the other hand, bottom-up linearization does not require action (i), leaving only the other two actions. Consequently, bottom-up linearization is shorter than top-down linearization. Figure~\ref{fig:parse} presents examples of the two linearization methods on binary and non-binary trees.

As illustrated in the figure, a special case arises for \textbf{bottom-up linearization of a non-binary tree}, which has not been studied in previous work: for each closing-nonterminal action ")", the start of the current constituent is unknown and hence needs to be predicted. Concretely, if action $a_k$ is predicted to be ")", we additionally predict the start position $s_k \in C_k$ given prefix $\boldsymbol{a}_{<k}$, where $C_k \subseteq \{1,\cdots,k-1\}$ is the set of feasible start positions, at each of which is either a token or ")" of a closed nonterminal that is not subsumed by any closed nonterminals yet at step $<k$. 
In the example of "\textit{Write an essay ) quickly )}", the feasible start position set of the first ")" is $\{1,2,3\}$. For position $i$, we concatenate the outputs of the first two attention heads in the final layer of the Transformer as its representation $\mathbf{h}_i$. We then compute $p(s_k\ | \ \boldsymbol{a}_{<k})$ as follows:
\[
p(s_{k} = i \ | \ \boldsymbol{a}_{<k}) \propto 
    \begin{cases}
    \exp{(\mathbf{h}_{k-1}^\top \Theta \mathbf{h}_{i})} & \text{if $i$ in $C_k$} \\
    0 & \text{otherwise}
    \end{cases}
\]
where $\Theta$ is an extra learnable matrix.
We then define $p(a_{k}\ | \ \boldsymbol{a}_{<k})$ as the product of probabilities of predicting ")" and the start position:
\[
p(a_{k}=(\text{")"},s_k)\ | \ \boldsymbol{a}_{<k}) =
    p(\text{")"}\ | \ \boldsymbol{a}_{<k}) \cdot p(s_k \ | \ \boldsymbol{a}_{<k})
\]

Another special case worthy of additional discussion is \textbf{top-down linearization of a binary tree}. For a binary tree, each constituent has a fixed length of two and therefore having both the opening-nonterminal and closing-nonterminal actions seems redundant. However, we decide to preserve both actions as explained in Appendix~\ref{app:bi-dn}.

\begin{figure}[tb]
    \centering
    \includegraphics[width=0.5\columnwidth]{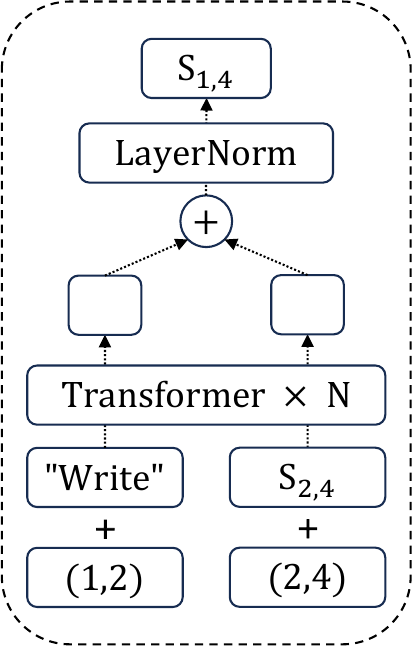}
    \caption{The Transformer-based external composition function $f$. The example input consists of the representations and position embeddings of two sub-constituents: "Write" and "an essay".} 
    \label{fig:composition}
\end{figure}

\subsection{Composition Function}\label{sec:composition}
When a constituent is completed, i.e., a ")" is generated, we use a composition function to compose all its sub-constituents into a single representation, which is then integrated into the Transformer and influences subsequent action generation. 
Previous studies use two different methods for this purpose: internal composition functions~\cite{sartran-etal-2022-Transformer} and external composition functions~\cite{yoshida-oseki-2022-composition, hu-etal-2024-generative}. 

An \textbf{internal} composition function, denoted as \textbf{\internal}, regards each composition as an additional action within the action sequence $\boldsymbol{a}$ and relies on the Transformer for composition computation. Specifically, for each predicted $a_k$ = ")", we directly input a ")" to the Transformer at step $k$ and set the attention mask such that only the sub-constituents of the current constituent can be attended to, thus forcing the computed hidden states to represent the composition of these sub-constituents. 
Note that step $k$ has an attention range limited to a single constituent and hence uninformative for predicting the next action. Therefore, we input a duplicate ")" to the Transformer at step $k+1$, allow attention to the full context (more details in section \ref{sec:masks}), and output the next action prediction. The duplicate ")" is then permanently masked in subsequent steps. 

The internal composition function, as described above, integrates the composition process into the Transformer, thus simplifying implementation and enabling parallelized training as in a standard Transformer LM. Its downside is that recursive composition through multiple Transformer layers has a receptive-field limitation as explained by section 2.3 of \citet{sartran-etal-2022-Transformer}) and the action sequence length is increased by the number of duplicate ")".

An \textbf{external} composition function, denoted as \textbf{\ext}, employs an additional module $f$ with separate parameters from the Transformer. Specifically, for each predicted $\alpha_k$ = ")", module $f$ takes as input the representations of the sub-constituents, which are either token embeddings or representations previously computed by the module, and outputs a single representation of the current constituent:
\[
    S_{p_0,p_m} = f(S_{p_0,p_1},\cdots,S_{p_{m-1},p_m})
\]
where $p_0, \ldots, p_m$ are left-inclusive and right-exclusive indexes of the sub-constituent spans.
The newly composed representation $S_{p_0,p_m}$ is then used as the input embedding at step 
$k$ in the Transformer. We adopt the Transformer-based composition function from GPST~\cite{hu-etal-2024-generative} shown in Figure~\ref{fig:composition}. 

The external composition function leverages the composed representation as the input to the Transformer, thereby avoiding the limitation of recursive composition in the internal composition function. 
However, it requires implementing and running an external module in addition to the Transformer. 
During training, all the compositions in the parse trees of the training set are typically pre-computed before the parallelized training of the Transformer, resulting in a slightly increased training time.

\begin{figure*}[tb]
\centering
\begin{subfigure}[t]{0.48\textwidth}
\centering
    \includegraphics[width=\textwidth,clip, trim=0.3cm 0.3cm 0cm 0.3cm]{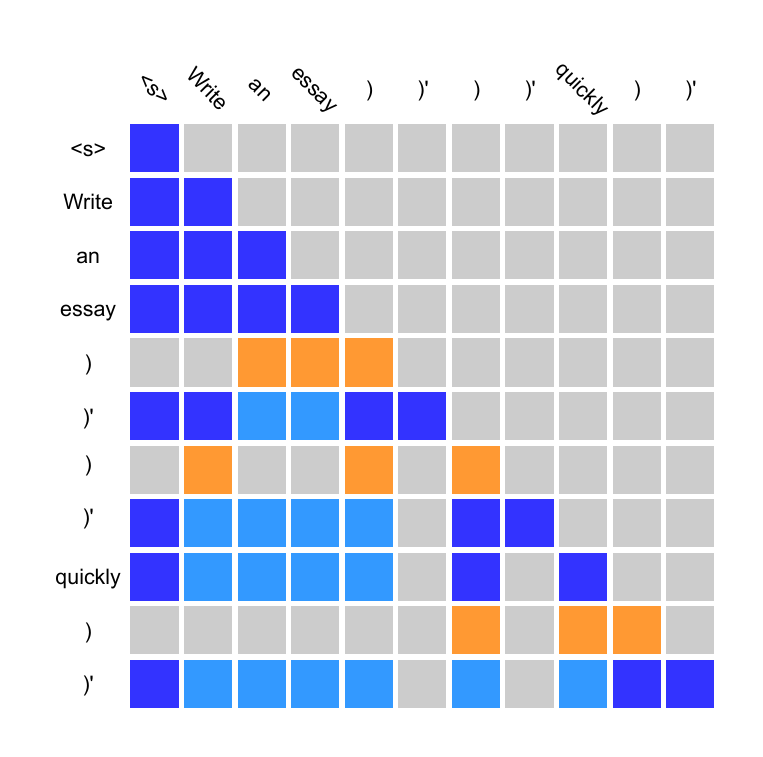} 
    \caption{Attention masks for modeling a binary tree with the internal composition function. ")'" represents a duplication of its preceding ")".}
    \label{fig:mask_int}
    \bigskip
\end{subfigure}
\hfill
\begin{subfigure}[t]{0.5\textwidth}
    \centering
    \includegraphics[width=\textwidth,clip, trim=0.35cm 0.35cm 0cm 0.35cm]{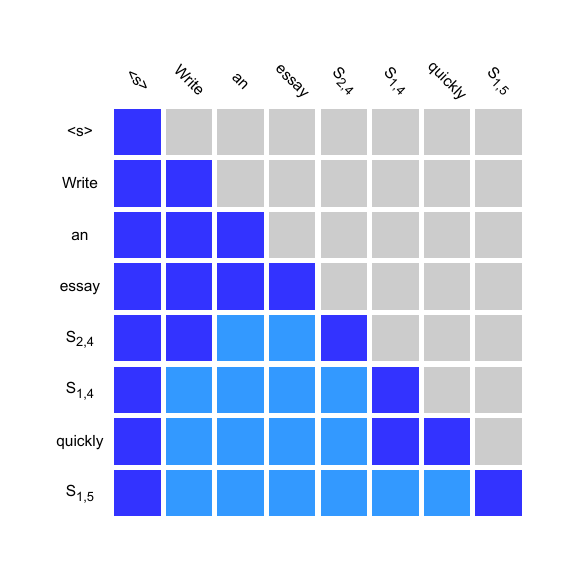} 
    \caption{Attention masks for modeling a binary tree with the external composition function. 
    $S_{2,4} = f(\text{an, essay})$, $S_{1,4} = f(\text{Write}, S_{2,4})$, $S_{1,5} = f(S_{1,4}, \text{quickly})$.}
    \label{fig:mask_ext}
\end{subfigure}
\caption{Examples of different mask patterns combined with different composition functions. We use \textcolor{gray}{gray} for masked positions, \textcolor{orange}{orange} for the attention ranges of internal compositions, \textcolor{dark-blue}{dark blue} for ordinary attended positions, \textcolor{light-blue}{light blue} for already composed positions that are only accessible in \textbf{Nm}.}
\label{fig:masks}
\end{figure*}

\subsection{Sub-Constituent Masking}\label{sec:masks}
After each composition, we may follow \citet{sartran-etal-2022-Transformer} and prevent subsequent steps from directly accessing information about already composed sub-constituents, creating a syntactic bottleneck that encourages learning informative compositions. On the other hand, from a language modeling perspective, allowing access to sub-constituent information, as done in \citet{hu-etal-2024-generative}, could enhance performance by providing additional context. Therefore, we consider two contrasting settings: (i) \textbf{Mask} the already composed sub-constituents, denoted as \textbf{\m}, and (ii) \textbf{No mask} for the already composed sub-constituents, denoted as \textbf{\nm}. An example illustrating different mask patterns combined with different composition functions is presented in Figure~\ref{fig:masks}. Note that the choice of \textbf{\m} or \textbf{\nm} does not affect the mask used for internal composition described in section~\ref{sec:composition}.

\subsection{Variants Within the Framework}\label{sec:variants}
Our framework specifies two options for each of the four key aspects and hence contains sixteen distinct SLMs, each named based on its configuration across the four aspects. For example, \textbf{\bi-\dn-\internal-\m} represents an SLM that models linearized binary trees in a top-down manner with an internal composition function and sub-constituent masking. We use the symbol \textbf{\#} to denote any option within a particular aspect. For instance, \textbf{\bi-\#-\#-\#} signifies an SLM that models binary trees, regardless of the choices made in the other aspects. 

Compositional SLMs from previous studies can be accommodated within our framework with minor modifications: (i) Transformer Grammars~\cite{sartran-etal-2022-Transformer} are classified as \textbf{\nb-\dn-\internal-\m} if modeling unlabeled trees. (ii) Composition Attention Grammars~\cite{yoshida-oseki-2022-composition} are classified as \textbf{\nb-\dn-\ext-\m} if we change the composition function from a bidirectional LSTM to a Transformer. (iii) Generative Pretrained Structured Transformers~\cite{hu-etal-2024-generative} are classified as \textbf{\bi-\up-\ext-\nm} if we set the depth of token layers to zero, i.e., we task type layers to predict both actions and tokens. Apart from these three models, the other thirteen SLMs within our framework are novel SLM variants not studied before.

\subsection{Inference} \label{sec:beam-search}

The space of token generation (type ii actions) is much larger than that of structure generation actions (type i \& iii actions) in SLMs, leading to an imbalance between their probabilities. Word-synchronous beam search, first introduced by \citet{stern-etal-2017-effective}, groups beams by the length of generated tokens instead of the whole action sequence, forcing SLMs to generate high-entropy tokens. We implement word-synchronous beam search for each SLM variant in our framework. There are two cases in which our implementation deviates from the standard implementation:
    (i) For \textbf{\nb-\#-\#-\#}, the number of nonterminals is not fixed given a sentence. We apply an additional hyperparameter $n_c$ as the maximum number of nonterminals, preventing models from composing too many times. 
    (ii) For \textbf{\#-\dn-\#-\#}, the model tends to generate a lot of successive "(" because structure generation is of low entropy. We also apply a hyperparameter $p_c$ as the maximum number of consecutive generation of "(".




\section{Experiments}

We compare the sixteen compositional SLMs from our framework with two Transformer baselines: (i) \textbf{GPT2-token}, a traditional language model of token sequences, and (ii) \textbf{GPT2-tree}, a syntactic language model of linearized trees without explicit composition. Following the setting in~\citet{sartran-etal-2022-Transformer}, GPT2-tree models non-binary trees in a top-down manner. We train all the models from scratch on the same corpus with comparable parameter sizes. We first evaluate all the models on language modeling and syntactic generalization. Then, we select eight best-performing compositional SLMs on both tasks, along with GPT2-token and GPT2-tree baselines, for further evaluation on summarization and dialogue---two generation tasks that we consider representative of downstream applications. Finally, we compare the inference efficiency of SLMs within the word-synchronous beam search setup. 


We also report experiments on additional baselines in Appendix~\ref{app:analysis}, including variants of GPT2-tree based on trivial trees and different binarization/linearization options.

\paragraph{Dataset and Preprocessing.} 
All the models are trained on the BLLIP{\small-LG} dataset of \citet{BLLIP-2000}, with training splits from \citet{hu-etal-2020-systematic}. We use an off-the-shelf CRF constituency parser~\cite{ijcai2020p0560} 
, implemented in \textit{Supar}\footnote{\url{https://github.com/yzhangcs/parser}}, to reparse the dataset and obtain silver constituency trees for training. All the silver trees parsed or sampled in the rest of the experiments are also produced with the same parser. Left-binarization is done with \textit{nltk}\footnote{\url{https://www.nltk.org/}}. Note that we model each sentence as a whole during both training and evaluation, and cutoff can only take place between two sentences to maintain the integrity of parse trees.

\paragraph{Hyper-parameters.} Following GPT-2$_\text{small}$~\cite{radford2019language}, we use 768-dimensional embeddings, a vocabulary size of 50257, 3072-dimensional hidden layer representations, 12 Transformer layers, and 12 attention heads for all SLMs and baselines. To maintain comparable parameter numbers between internal and external composition functions, we use a relatively small Transformer as the external composition function for \textbf{\#-\#-\ext-\#}, setting the input dimension to 256 and the number of layers to 4, following~\cite{hu-etal-2024-generative}. Token embeddings are down-scaled before composition and the constituent representations are up-scaled before fed into the main SLM Transformer. The module increases the total parameter number by only 5\%, which we believe does not significantly affect the comparability between models. We discuss other training details and training variances in Appendix~\ref{app:variance}.

\subsection{Document-Level Language Modeling}
\paragraph{Dataset.} We evaluate all the models on the testing split of BLLIP{\small-LG} from \citet{hu-etal-2020-systematic}.
\paragraph{Setup.} Since SLMs model $p(\mathbf{x}, \mathbf{y})$, the joint probability of sentences and parse trees, we compute the probability of a sentence as $p(\mathbf{x}) = \sum_\mathbf{y} p(\mathbf{x}, \mathbf{y})$. It is impossible to compute the summation exactly due to the large space of possible constituency trees, so we follow \citet{sartran-etal-2022-Transformer} to approximate it using a relatively small set of trees sampled from a proposal model. We use a CRF parser as the proposal model and sample 300 unlabeled constituency trees without replacement as a proposal tree set $\mathbf{Y'}$. $p(\mathbf{x})$ is then approximated by $\sum_{\mathbf{y} \in \mathbf{Y'}}p(\mathbf{x},\mathbf{y})$, which is a lower bound of the true value (hence leading to an upper bound of perplexity). 

For document-level language modeling, we compute the probability of a document consisting of $M$ sentences.
When computing $p(\mathbf{x}^{i}|\mathbf{x}^{0}, \cdots,\mathbf{x}^{i-1})$, the probability of the $i$-th sentence in the document conditioned on its $i-1$ preceding sentences, in theory we have to marginalize over all the $i$ parse trees, each having 300 samples, which demands unacceptable computational costs. Following \citet{sartran-etal-2022-Transformer}, we approximate this by greedily choosing a single tree $\mathbf{y}$ that maximizes $p(\mathbf{x}, \mathbf{y})$ for each of the preceding $i-1$ sentences, serving as a single-path prefix for the $i$-th sentence. 
\begin{table}[tb]
  \centering
  \small
  \begin{tabular}{lcc} \toprule
    \thead{Model}           & \thead{PPL$^\dagger$ ($\downarrow$)} & \thead{SG ($\uparrow$)} \\\midrule
    \text{GPT2-token} & \cellcolor[rgb]{0.0, 0.8, 0.0}{\textbf{17.31}} & \cellcolor[rgb]{0.8784313725490196, 0.9756862745098039, 0.8784313725490196}{64.1} \\
    \text{GPT2-tree} & \cellcolor[rgb]{0.5333333333333333, 0.9066666666666667, 0.5333333333333333}{19.97} & \cellcolor[rgb]{0.44705882352941173, 0.8894117647058823, 0.44705882352941173}{73.1} \\
    {\bi-\up-\ext-\nm} & \cellcolor[rgb]{0.6431372549019607, 0.9286274509803921, 0.6431372549019607}{20.51} & \cellcolor[rgb]{0.1098039215686275, 0.8219607843137255, 0.1098039215686275}{80.1} \\
    {\bi-\up-\ext-\m} & \cellcolor[rgb]{1.0, 0.6956862745098039, 0.6956862745098039}{24.15} & \cellcolor[rgb]{0.0, 0.8, 0.0}{\textbf{82.4}} \\
    {\bi-\up-\internal-\nm} & \cellcolor[rgb]{0.5411764705882353, 0.908235294117647, 0.5411764705882353}{19.99} & \cellcolor[rgb]{0.23529411764705888, 0.8470588235294119, 0.23529411764705888}{77.5} \\
    {\bi-\up-\internal-\m} & \cellcolor[rgb]{0.807843137254902, 0.9615686274509804, 0.807843137254902}{21.32} & \cellcolor[rgb]{0.12549019607843137, 0.8250980392156864, 0.12549019607843137}{79.7} \\
    {\bi-\dn-\ext-\nm} & \cellcolor[rgb]{1.0, 0.7772549019607843, 0.7772549019607843}{23.62} & \cellcolor[rgb]{0.10196078431372546, 0.8203921568627451, 0.10196078431372546}{80.2} \\
    {\bi-\dn-\ext-\m} & \cellcolor[rgb]{1.0,0.2,0.2}{27.21} & \cellcolor[rgb]{0.07058823529411762, 0.8141176470588236, 0.07058823529411762}{80.9} \\
    {\bi-\dn-\internal-\nm} & \cellcolor[rgb]{0.9490196078431372, 0.9898039215686274, 0.9490196078431372}{22.02} & \cellcolor[rgb]{0.14117647058823535, 0.8282352941176471, 0.14117647058823535}{79.4} \\
    {\bi-\dn-\internal-\m} & \cellcolor[rgb]{1.0, 0.3129411764705883, 0.3129411764705883}{26.50} & \cellcolor[rgb]{0.07058823529411762, 0.8141176470588236, 0.07058823529411762}{80.9} \\
    {\nb-\up-\ext-\nm} & \cellcolor[rgb]{1.0, 0.7396078431372548, 0.7396078431372548}{23.85} & \cellcolor[rgb]{1.0,0.2,0.2}{40.8} \\
    {\nb-\up-\ext-\m} & \cellcolor[rgb]{1.0, 0.7082352941176471, 0.7082352941176471}{24.07} & \cellcolor[rgb]{1.0, 0.6203921568627451, 0.6203921568627451}{51.8} \\
    {\nb-\up-\internal-\nm} & \cellcolor[rgb]{0.4156862745098039, 0.8831372549019608, 0.4156862745098039}{19.36} & \cellcolor[rgb]{0.1333333333333333, 0.8266666666666667, 0.1333333333333333}{79.6} \\
    {\nb-\up-\internal-\m} & \cellcolor[rgb]{0.9490196078431372, 0.9898039215686274, 0.9490196078431372}{22.01} & \cellcolor[rgb]{0.43137254901960786, 0.8862745098039216, 0.43137254901960786}{73.4} \\
    {\nb-\dn-\ext-\nm} & \cellcolor[rgb]{0.7215686274509804, 0.9443137254901961, 0.7215686274509804}{20.88} & \cellcolor[rgb]{1.0, 0.5952941176470589, 0.5952941176470589}{51.1} \\
    {\nb-\dn-\ext-\m} & \cellcolor[rgb]{1.0, 0.5325490196078431, 0.5325490196078431}{25.15} & \cellcolor[rgb]{1.0, 0.6266666666666667, 0.6266666666666667}{51.9} \\
    {\nb-\dn-\internal-\nm} & \cellcolor[rgb]{0.1568627450980392, 0.8313725490196079, 0.1568627450980392}{18.11} & \cellcolor[rgb]{0.20392156862745103, 0.8407843137254902, 0.20392156862745103}{78.1} \\
    {\nb-\dn-\internal-\m} & \cellcolor[rgb]{1.0, 0.9905882352941177, 0.9905882352941177}{22.30} & \cellcolor[rgb]{0.32156862745098036, 0.8643137254901961, 0.32156862745098036}{75.6} \\ \bottomrule
  \end{tabular}
  \caption{\label{citation-guide}
   Perplexity (PPL) and syntactic generalization (SG) results of our models and baselines. $^\dagger$: All the reported PPLs except that of GPT2-token are upper bounds of the true values.
  }
  \label{table:sg-ppl}
\end{table}

\paragraph{Results.}
We report the perplexity of all the models in Table~\ref{table:sg-ppl}. All the SLMs, including GPT2-tree, show higher perplexity than GPT2-token baselines, seemingly implying that document-level language modeling may not benefit from the inductive bias of syntax. However, this observation is inconclusive because the reported SLM PPLs are upper bounds of the true values. 
Another observation is that only \textbf{\nb-\dn-\internal-\nm} and \textbf{\nb-\up-\internal-\nm} outperform GPT2-tree, and \textbf{\bi-\up-\internal-\nm} shows comparable performance with GPT2-tree. Since the main difference between GPT2-tree and our models is that it does not involve explicit composition, this observation indicates that explicit composition may not be critical for language modeling and only helps in certain configurations. 

Comparing compositional SLMs in our framework, we have two major findings: (i) Fixing the first three aspects, \textbf{\#-\#-\#-\nm} consistently shows significantly better language modeling performance than \textbf{\#-\#-\#-\m}, which is to be expected because less information is directly available at each generation step in the setting of \textbf{\m}, making it harder for next token prediction. (ii) \textbf{\#-\#-\internal-\#} achieves lower perplexity than \textbf{\#-\#-\ext-\#}, showing that directly reusing parameters of the main Transformer for composition is a better choice for language modeling than using a small external composition function. 

\begin{table*}[ht]
  \centering
  \small
  \begin{tabular}{l|cccc|cccc} \toprule
    \makecell[c]{\multirow{3}{*}{\thead{Model}}} & \multicolumn{4}{c|}{\thead{Xsum}} & \multicolumn{4}{c}{\thead{DailyDialog}} \\ \cmidrule{2-9}
    & \thead{R-1} & \thead{R-2} & \thead{R-L} & \thead{R-AVG} & \thead{R-1} & \thead{R-2} & \thead{R-L} & \thead{R-AVG}  \\
    \midrule
    \text{GPT2-token} & \cellcolor[rgb]{0.8, 0.96, 0.8}{27.14} & \cellcolor[rgb]{0.9647058823529414, 0.9929411764705883, 0.9647058823529414}{7.67} & \cellcolor[rgb]{0.8156862745098039, 0.9631372549019608, 0.8156862745098039}{21.65} & \cellcolor[rgb]{0.8470588235294118, 0.9694117647058823, 0.8470588235294118}{18.82} & \cellcolor[rgb]{0.6274509803921569, 0.9254901960784314, 0.6274509803921569}14.02 & \cellcolor[rgb]{0.3058823529411765, 0.8611764705882353, 0.3058823529411765}3.82 & \cellcolor[rgb]{0.6666666666666667, 0.9333333333333333, 0.6666666666666667}13.31 & \cellcolor[rgb]{0.5647058823529412, 0.9129411764705883, 0.5647058823529412}10.38\\
    \text{GPT2-tree} & \cellcolor[rgb]{0.0, 0.8, 0.0}{\textbf{29.59}} & \cellcolor[rgb]{0.0, 0.8, 0.0}{\textbf{9.47}} & \cellcolor[rgb]{0.0, 0.8, 0.0}{\textbf{23.58}} & \cellcolor[rgb]{0.0, 0.8, 0.0}{\textbf{20.88}} & \cellcolor[rgb]{0.0, 0.8, 0.0}\textbf{14.99} & \cellcolor[rgb]{0.2901960784313725, 0.8580392156862745, 0.2901960784313725}3.83 & \cellcolor[rgb]{0.0, 0.8, 0.0}\textbf{14.31} & \cellcolor[rgb]{0.0, 0.8, 0.0}\textbf{11.04}\\
    {\bi-\up-\ext-\nm} & \cellcolor[rgb]{0.18039215686274512, 0.8360784313725491, 0.18039215686274512}{29.04} & \cellcolor[rgb]{0.27450980392156865, 0.8549019607843138, 0.27450980392156865}{8.95} & \cellcolor[rgb]{0.23529411764705888, 0.8470588235294119, 0.23529411764705888}{23.01} & \cellcolor[rgb]{0.2196078431372549, 0.843921568627451, 0.2196078431372549}{20.33} & \cellcolor[rgb]{0.5490196078431373, 0.9098039215686274, 0.5490196078431373}14.14 & \cellcolor[rgb]{0.0, 0.8, 0.0}\textbf{4.01} & \cellcolor[rgb]{0.47058823529411764, 0.8941176470588236, 0.47058823529411764}13.61 & \cellcolor[rgb]{0.38431372549019605, 0.8768627450980393, 0.38431372549019605}10.59 \\
    {\bi-\up-\ext-\m} & \cellcolor[rgb]{1.0,0.2,0.2}{23.48} & \cellcolor[rgb]{1.0,0.2,0.2}{5.75} & \cellcolor[rgb]{1.0,0.2,0.2}{18.84} & \cellcolor[rgb]{1.0,0,0}{16.02} & \cellcolor[rgb]{1.0, 0.4698039215686275, 0.4698039215686275}12.44 & \cellcolor[rgb]{1.0, 0.4823529411764706, 0.4823529411764706}3.00 & \cellcolor[rgb]{1.0, 0.38823529411764707, 0.38823529411764707}11.69 & \cellcolor[rgb]{1.0, 0.41960784313725497, 0.41960784313725497}9.04\\
    {\bi-\up-\internal-\nm} & \cellcolor[rgb]{0.2117647058823532, 0.8423529411764706, 0.2117647058823532}{28.93} & \cellcolor[rgb]{0.2666666666666667, 0.8533333333333334, 0.2666666666666667}{8.97} & \cellcolor[rgb]{0.25098039215686274, 0.8501960784313726, 0.25098039215686274}{22.97} & \cellcolor[rgb]{0.2431372549019608, 0.8486274509803922, 0.2431372549019608}{20.29} & \cellcolor[rgb]{1.0, 0.7647058823529411, 0.7647058823529411}13.01 & \cellcolor[rgb]{1.0, 0.9152941176470588, 0.9152941176470588}3.33 & \cellcolor[rgb]{1.0, 0.6580392156862744, 0.6580392156862744}12.19 & \cellcolor[rgb]{1.0, 0.7458823529411764, 0.7458823529411764}9.51\\
    {\bi-\up-\internal-\m} & \cellcolor[rgb]{1.0, 0.5513725490196079, 0.5513725490196079}{24.84} & \cellcolor[rgb]{1.0, 0.5827450980392157, 0.5827450980392157}{6.64} & \cellcolor[rgb]{1.0, 0.5513725490196079, 0.5513725490196079}{19.88} & \cellcolor[rgb]{1.0, 0.5576470588235294, 0.5576470588235294}{17.12} & \cellcolor[rgb]{1.0, 0.2627450980392157, 0.2627450980392157}12.05 & \cellcolor[rgb]{1.0, 0.2, 0.2}2.78 & \cellcolor[rgb]{1.0, 0.23764705882352943, 0.23764705882352943}11.40 & \cellcolor[rgb]{1.0, 0.21254901960784314, 0.21254901960784314}8.74\\
    {\nb-\up-\internal-\nm} & \cellcolor[rgb]{0.1725490196078432, 0.8345098039215687, 0.1725490196078432}{29.05} & \cellcolor[rgb]{0.2196078431372549, 0.843921568627451, 0.2196078431372549}{9.06} & \cellcolor[rgb]{0.14901960784313728, 0.8298039215686275, 0.14901960784313728}{23.21} & \cellcolor[rgb]{0.18039215686274512, 0.8360784313725491, 0.18039215686274512}{20.44} & \cellcolor[rgb]{0.7686274509803921, 0.9537254901960784, 0.7686274509803921}13.81 & \cellcolor[rgb]{0.5333333333333333, 0.9066666666666667, 0.5333333333333333}3.68 & \cellcolor[rgb]{0.8156862745098039, 0.9631372549019608, 0.8156862745098039}13.09 & \cellcolor[rgb]{0.7294117647058824, 0.9458823529411765, 0.7294117647058824}10.19\\
    {\nb-\up-\internal-\m} & \cellcolor[rgb]{1.0, 0.41333333333333333, 0.41333333333333333}{24.30} & \cellcolor[rgb]{1.0, 0.4258823529411765, 0.4258823529411765}{6.28} & \cellcolor[rgb]{1.0, 0.4007843137254902, 0.4007843137254902}{19.45} & \cellcolor[rgb]{1.0, 0.41333333333333333, 0.41333333333333333}{16.68} & \cellcolor[rgb]{1.0, 0.2, 0.2}11.92 & \cellcolor[rgb]{1.0, 0.39450980392156865, 0.39450980392156865}2.93 & \cellcolor[rgb]{1.0, 0.2, 0.2}11.33 & \cellcolor[rgb]{1.0, 0.2, 0.2}8.72\\
    {\nb-\dn-\internal-\nm} & \cellcolor[rgb]{0.03137254901960784, 0.8062745098039217, 0.03137254901960784}{29.48} & \cellcolor[rgb]{0.03137254901960784, 0.8062745098039217, 0.03137254901960784}{9.40} & \cellcolor[rgb]{0.015686274509803977, 0.8031372549019609, 0.015686274509803977}{23.54} & \cellcolor[rgb]{0.02352941176470591, 0.8047058823529412, 0.02352941176470591}{20.81} & \cellcolor[rgb]{0.6509803921568628, 0.9301960784313725, 0.6509803921568628}13.99 & \cellcolor[rgb]{0.2588235294117647, 0.851764705882353, 0.2588235294117647}3.85 & \cellcolor[rgb]{0.6352941176470588, 0.9270588235294118, 0.6352941176470588}13.36 & \cellcolor[rgb]{0.5490196078431373, 0.9098039215686274, 0.5490196078431373}10.40\\
    {\nb-\dn-\internal-\m} & \cellcolor[rgb]{1.0, 0.8839215686274511, 0.8839215686274511}{26.10} & \cellcolor[rgb]{1.0, 0.8713725490196078, 0.8713725490196078}{7.31} & \cellcolor[rgb]{1.0, 0.9215686274509804, 0.9215686274509804}{20.98} & \cellcolor[rgb]{1.0, 0.8713725490196078, 0.8713725490196078}{18.07} & \cellcolor[rgb]{1.0, 0.5011764705882353, 0.5011764705882353}12.50 & \cellcolor[rgb]{1.0, 0.8901960784313727, 0.8901960784313727}3.31 & \cellcolor[rgb]{1.0, 0.5011764705882353, 0.5011764705882353}11.90 & \cellcolor[rgb]{1.0, 0.5513725490196079, 0.5513725490196079}9.23\\
    \bottomrule
  \end{tabular}
  \caption{Results on the summarization and dialogue tasks.}
  \label{table:xsum}
\end{table*}
\subsection{Syntactic Generalization}\label{exp:sg}
\paragraph{Dataset.}
We evaluate all the models on the syntactic generalization (SG) task~\cite{hu-etal-2020-systematic}, consisting of test suites for six fine-grained syntactic phenomena. 
\paragraph{Setup.}
Each test suite is evaluated by a specific inequality formula, which requires computing the surprisal values, i.e., $-\log p(\mathbf{x}_t|\mathbf{x}_{<t})$. We compute the surprisal values for SLMs using the word-synchronous beam search described in section~\ref{sec:beam-search}. As the target token $\mathbf{x}_t$ is given, we modify the algorithm by directly predicting the given token. 
The beam size is set to 300. The maximum number of nonterminals $n_c$ is dynamically set to the length of each sentence and the maximum number of consecutive opening-nonterminal actions $p_c$ is set to 3. Further details on the selection of these hyperparameters are provided in Appendix \ref{app:hyperparameters}.
\paragraph{Results.}
The overall results are reported in Table~\ref{table:sg-ppl}. We also plot the detailed performance over each of the six syntactic phenomena in Appendix~\ref{app:sg}. Most of the SLMs outperform the GPT2-token baseline with a significant gain in the SG score, which is to be expected because of their explicit modeling of syntax. 
Furthermore, most of the compositional SLMs outperform GPT2-tree, proving that explicit composition is helpful to SLMs in syntactic modeling.

It is notable that four compositional SLMs have extremely low SG scores and they all belong to the configuration of \textbf{\nb-\#-\ext-\#}, i.e., modeling non-binary trees with an external composition function. This is likely because the relatively small external composition model fails to capture the complicated interactions among varying numbers of sub-constituents. 
In sharp contrast, external composition functions applied to binary trees (\textbf{\bi-\#-\ext-\#}) achieve impressive SG scores, occupying four of the top five spots, suggesting that they are expressive enough to handle binary composition and even outperform internal composition functions of much larger sizes. 
A similar trend can be observed on internal composition functions regarding the relative difficulty of modeling non-binary composition in comparison with binary composition (i.e., \textbf{\nb-\#-\internal-\#} vs. \textbf{\bi-\#-\internal-\#}), although to a much lesser extent.

For sub-constituent masking, we observe that \textbf{\bi-\#-\#-\m} always performs better than \textbf{\bi-\#-\#-\nm}, confirming that the information bottleneck created by sub-constituent masking can benefit syntactic modeling. On the other hand, when it comes to non-binary trees (excluding worst-performing \textbf{\nb-\#-\ext-\#}), we observe that \textbf{\nb-\#-\internal-\m} performs worse than \textbf{\nb-\#-\internal-\nm}. Considering that non-binary composition is more difficult as discussed above, we may conclude that sub-constituent masking is useful to syntactic modeling only when composition is effective.

\subsection{Downstream Tasks}
\subsubsection{Summarization}\label{exp:xsum}
\paragraph{Dataset.} 
We conduct experiments on the BBC extreme dataset (Xsum)~\cite{narayan-etal-2018-dont} to assess the performance of SLMs in terms of summarization abilities. 

\paragraph{Setup.}
We truncate the documents and their summaries to 600 and 70, respectively, and concatenate them with a short prompt "\textit{Summary:}". Following \citet{hu-etal-2024-generative}, we finetune each model for 15 epochs with a batch size of 16 on the training split of Xsum. ROUGE~\cite{lin-hovy-2003-automatic} is employed as the evaluation metric. To evaluate SLMs, we apply the word-synchronous beam search to top-$k$ random sampling with $k$ set to 2. 
The maximum number of nonterminals $n_c$ is dynamically set to the length of each sentence and the maximum number of consecutive opening-nonterminal actions $p_c$ is set to 5. For all the SLMs, the input contains the linearization of the sentences in the document and their corresponding silver parse trees.

We only conduct experiments on eight compositional SLMs and discard the other eight as explained below:
(i) The four models of \textbf{\nb-\#-\ext-\#} show poor performance on language modeling and syntactic generalization, indicating a failure in composition learning.
(ii) The four models of \textbf{\bi-\dn-\#-\#} model both the opening-nonterminal and the closing-nonterminal actions. For binary trees, these two actions are redundant, and predicting one of them for each constituent is enough (as done in \textbf{\bi-\up-\#-\#}). The four models also show poor language modeling performance. 

\paragraph{Results.}
The results are presented in Table~\ref{table:xsum}. 
First of all, \textbf{\#-\#-\#-\nm} significantly outperforms \textbf{\#-\#-\#-\m}, which is consistent with the language modeling results and highlights the importance of direct access to composed sub-constituents in generation tasks.

Second, all the four SLMs of \textbf{\#-\#-\#-\nm} outperform GPT2-token, which can be attributed to two possible reasons: 
(i) SLMs may have better generation abilities than GPT2-token. (ii) GPT2-token only receives the input text as the prompt, while SLMs receive additional information---linearized parse trees of the input text. 
Regardless of the reasons, The results suggest that SLMs aided by an off-the-shelf parser have great potential in downstream generation tasks. 

Finally, GPT2-tree achieves the best scores while compositional SLMs of \textbf{\#-\#-\#-\nm} show comparable or slightly lower scores, suggesting again that explicit composition is not critical in generation tasks. 

\subsubsection{Dialogue}
\paragraph{Dataset.} 
We conduct experiments on the Dailydialog dataset~\cite{li-etal-2017-dailydialog} to assess the performance of SLMs in terms of dialogue abilities.

\paragraph{Setup.}
In each dialogue, all the utterances except for the last one are fed as the prompt, and the last utterance is the generation target. We add a special \texttt{<sep>} to indicate the start of each utterance. We truncate the prompt utterances and the target to 600 and 150, respectively. We finetune each model for 5 epochs with a batch size of 16 on the training split of Dailydialog. The other setups are exactly the same as in the summarization experiments.

\paragraph{Results.}
The results are presented in Table~\ref{table:xsum}. 
Similar to the findings in the summarization task, \textbf{\#-\#-\#-\nm} significantly outperforms \textbf{\#-\#-\#-\m} and GPT2-tree achieves the best scores. Different from the summarization task, however, only \textbf{\bi-\up-\ext-\nm} clearly outperforms GPT2-token among the four \textbf{\#-\#-\#-\nm} SLMs. 
Nonetheless, the improved performance of GPT2-tree and \textbf{\bi-\up-\ext-\nm} still shows the potential of SLMs in downstream generation tasks.

\begin{table}
  \centering
  \small
  \begin{tabular}{lcccc}
    \toprule
    \textbf{Model}           & \textbf{bsz-10} & \textbf{bsz-30} & \textbf{bsz-100} & \textbf{bsz-300} \\
    \midrule
    \text{GPT2-tree} & \cellcolor[rgb]{0.5882352941176471, 0.9176470588235295, 0.5882352941176471}{2.06} & \cellcolor[rgb]{0.8431372549019608, 0.9686274509803922, 0.8431372549019608}{2.94} & \cellcolor[rgb]{0.9215686274509803, 0.9843137254901961, 0.9215686274509803}{3.98} & \cellcolor[rgb]{0.8627450980392157, 0.9725490196078431, 0.8627450980392157}{5.81} \\
    {\bi-\up-\ext-\nm} & \cellcolor[rgb]{0.17647058823529413, 0.8352941176470589, 0.17647058823529413}{1.28} & \cellcolor[rgb]{0.09803921568627451, 0.819607843137255, 0.09803921568627451}{1.46} & \cellcolor[rgb]{0.0196078431372549, 0.803921568627451, 0.0196078431372549}{1.77} & \cellcolor[rgb]{0.0, 0.8, 0.0}{2.62} \\
    {\bi-\up-\ext-\m} & \cellcolor[rgb]{0.19607843137254902, 0.8392156862745098, 0.19607843137254902}{1.33} & \cellcolor[rgb]{0.11764705882352941, 0.823529411764706, 0.11764705882352941}{1.49} & \cellcolor[rgb]{0.09803921568627451, 0.819607843137255, 0.09803921568627451}{1.97} & \cellcolor[rgb]{0.1568627450980392, 0.8313725490196079, 0.1568627450980392}{3.23} \\
    {\bi-\up-\internal-\nm} & \cellcolor[rgb]{1.0, 0.8509803921568627, 0.8509803921568627}{4.28} & \cellcolor[rgb]{1.0, 0.8392156862745098, 0.8392156862745098}{4.81} & \cellcolor[rgb]{1.0, 0.8549019607843137, 0.8549019607843137}{5.95} & \cellcolor[rgb]{1.0, 0.8666666666666667, 0.8666666666666667}{8.69} \\
    {\bi-\up-\internal-\m} & \cellcolor[rgb]{1.0, 0.8509803921568627, 0.8509803921568627}{4.28} & \cellcolor[rgb]{1.0, 0.8392156862745098, 0.8392156862745098}{4.80} & \cellcolor[rgb]{1.0, 0.8588235294117647, 0.8588235294117647}{5.91} & \cellcolor[rgb]{1.0, 0.8666666666666667, 0.8666666666666667}{8.70} \\
    {\bi-\dn-\ext-\nm} & \cellcolor[rgb]{0.13725490196078433, 0.8274509803921569, 0.13725490196078433}{1.20} & \cellcolor[rgb]{0.0392156862745098, 0.807843137254902, 0.0392156862745098}{1.33} & \cellcolor[rgb]{0.1568627450980392, 0.8313725490196079, 0.1568627450980392}{2.13} & \cellcolor[rgb]{0.3137254901960784, 0.8627450980392157, 0.3137254901960784}{3.76} \\
    {\bi-\dn-\ext-\m} & \cellcolor[rgb]{0.09803921568627451, 0.819607843137255, 0.09803921568627451}{1.14} & \cellcolor[rgb]{0.17647058823529413, 0.8352941176470589, 0.17647058823529413}{1.61} & \cellcolor[rgb]{0.29411764705882354, 0.8588235294117648, 0.29411764705882354}{2.44} & \cellcolor[rgb]{0.5098039215686274, 0.9019607843137255, 0.5098039215686274}{4.53} \\
    {\bi-\dn-\internal-\nm} & \cellcolor[rgb]{1.0, 0.9019607843137255, 0.9019607843137255}{3.79} & \cellcolor[rgb]{1.0, 0.8745098039215686, 0.8745098039215686}{4.46} & \cellcolor[rgb]{1.0, 0.8470588235294119, 0.8470588235294119}{6.03} & \cellcolor[rgb]{1.0, 0.8117647058823529, 0.8117647058823529}{9.73} \\
    {\bi-\dn-\internal-\m} & \cellcolor[rgb]{1.0, 0.8901960784313725, 0.8901960784313725}{3.91} & \cellcolor[rgb]{1.0, 0.8784313725490196, 0.8784313725490196}{4.44} & \cellcolor[rgb]{1.0, 0.8549019607843137, 0.8549019607843137}{5.93} & \cellcolor[rgb]{1.0, 0.7725490196078432, 0.7725490196078432}{10.43} \\
    {\nb-\up-\ext-\nm} & \cellcolor[rgb]{0.21568627450980393, 0.8431372549019608, 0.21568627450980393}{1.35} & \cellcolor[rgb]{0.29411764705882354, 0.8588235294117648, 0.29411764705882354}{1.83} & \cellcolor[rgb]{0.3333333333333333, 0.8666666666666667, 0.3333333333333333}{2.53} & \cellcolor[rgb]{0.5098039215686274, 0.9019607843137255, 0.5098039215686274}{4.53} \\
    {\nb-\up-\ext-\m} & \cellcolor[rgb]{1.0, 0.9803921568627451, 0.9803921568627451}{3.04} & \cellcolor[rgb]{1.0, 0.9725490196078431, 0.9725490196078431}{3.51} & \cellcolor[rgb]{1.0, 0.9568627450980393, 0.9568627450980393}{4.71} & \cellcolor[rgb]{1.0, 0.9215686274509804, 0.9215686274509804}{7.74} \\
    {\nb-\up-\internal-\nm} & \cellcolor[rgb]{1.0, 0.4627450980392157, 0.4627450980392157}{7.98} & \cellcolor[rgb]{1.0, 0.4235294117647058, 0.4235294117647058}{8.90} & \cellcolor[rgb]{1.0, 0.4235294117647058, 0.4235294117647058}{11.23} & \cellcolor[rgb]{1.0, 0.4117647058823529, 0.4117647058823529}{16.97} \\
    {\nb-\up-\internal-\m} & \cellcolor[rgb]{1.0, 0.2, 0.2}{10.52} & \cellcolor[rgb]{1.0, 0.2, 0.2}{11.12} & \cellcolor[rgb]{1.0, 0.2, 0.2}{13.97} & \cellcolor[rgb]{1.0, 0.2, 0.2}{20.84} \\
    {\nb-\dn-\ext-\nm} & \cellcolor[rgb]{0.0, 0.8, 0.0}{0.93} & \cellcolor[rgb]{0.0, 0.8, 0.0}{1.25} & \cellcolor[rgb]{0.0, 0.8, 0.0}{1.70} & \cellcolor[rgb]{0.19607843137254902, 0.8392156862745098, 0.19607843137254902}{3.35} \\
    {\nb-\dn-\ext-\m} & \cellcolor[rgb]{0.0784313725490196, 0.8156862745098039, 0.0784313725490196}{1.11} & \cellcolor[rgb]{0.21568627450980393, 0.8431372549019608, 0.21568627450980393}{1.71} & \cellcolor[rgb]{0.39215686274509803, 0.8784313725490196, 0.39215686274509803}{2.66} & \cellcolor[rgb]{0.6862745098039216, 0.9372549019607843, 0.6862745098039216}{5.17} \\
    {\nb-\dn-\internal-\nm} & \cellcolor[rgb]{1.0, 0.8823529411764706, 0.8823529411764706}{3.99} & \cellcolor[rgb]{1.0, 0.8156862745098039, 0.8156862745098039}{5.03} & \cellcolor[rgb]{1.0, 0.7490196078431373, 0.7490196078431373}{7.22} & \cellcolor[rgb]{1.0, 0.7215686274509805, 0.7215686274509805}{11.37} \\
    {\nb-\dn-\internal-\m} & \cellcolor[rgb]{1.0, 0.8352941176470589, 0.8352941176470589}{4.43} & \cellcolor[rgb]{1.0, 0.7843137254901962, 0.7843137254901962}{5.34} & \cellcolor[rgb]{1.0, 0.7098039215686274, 0.7098039215686274}{7.72} & \cellcolor[rgb]{1.0, 0.6745098039215687, 0.6745098039215687}{12.17} \\
    \bottomrule
  \end{tabular}
  \caption{
   Inference time (in seconds, lower is better). "\textbf{bsz}" refers to beam size.
  }
  \label{table:beam}
\end{table}

\subsection{Inference Efficiency}\label{exp:eff}
\paragraph{Setup.}
In practice, SLMs reply on word-synchronous beam search to generate meaningful sentences and proper tree structures. Therefore, we compare the inference efficiency of all the SLMs with word-synchronous beam search. GPT2-token is not considered here because it only generates tokens without the need for synchronization of structures (see Appendix~\ref{app:eff} for additional experiments comparing the efficiency gap between GPT2-token with SLMs). We follow the same setup in the syntactic generalization experiment. The sentence is fixed in advance, ensuring a fair comparison between the inference time of different SLMs. We evaluate the SLMs on five sentences of 20 tokens each. For beam sizes of \{10, 30, 100, 300\}, we repeat the inference for 5 times and compute the average time for each. We also choose one sentence and count the number of model forward calls for each beam size as a supplement in Appendix~\ref{app:call}. 
All the efficiency evaluation experiments are run on a single H800 GPU unless specifically mentioned. 
\paragraph{Results.}
The results are shown in Table~\ref{table:beam}. All the models with external composition functions (\textbf{\ext}) show much less inference time than those with internal ones (\textbf{\internal}), because the external composition module is smaller and faster than reusing the main Transformer for composition. Moreover, a large gap in inference time exists between \textbf{\#-\#-\ext-\m} and \textbf{\#-\#-\ext-\nm}. This is because when \textbf{\nm} is combined with \textbf{\ext}, the attention mask becomes a simple casual mask which can be accelerated by various optimization methods in \texttt{scaled\_dot\_product\_attention} of PyTorch. 

The \textbf{\nb-\up-\#-\#} models require more time than the \textbf{\nb-\dn-\#-\#} models, despite having shorter action sequences.
We also find that \textbf{\bi-\#-\#-\#} is generally faster than \textbf{\nb-\#-\#-\#}, which is quite surprising because a non-binary tree usually has fewer nonterminals than a binary tree and thus a shorter linearization. We speculate that these two observations result from an intrinsic problem of applying word-synchronous beam search to modeling linearized non-binary trees, which we discuss further in Appendix~\ref{app:inference}. 


\subsection{Overall Observations}
Based on the overall experimental results, we recommend several design choices for compositional SLMs:

(i) SLMs without sub-constituent masks generally outperform their masked counterparts in both effectiveness and efficiency except for a potentially small disadvantage in syntax-focused tasks.

(ii) External composition excels for efficiency. If efficiency is the main concern, modeling binary trees with an external composition function is a good choice with decent performance on all the tasks.

(iii) Binary trees align better with bottom-up linearization in terms of both efficiency and performance, while non-binary trees seem to be more compatible with top-down linearization.

(iv) Modeling non-binary trees using an external composition function can result in suboptimal performance on certain tasks.

\section{Related Work}
Augmenting language models with syntactic bias has been a longstanding area of research. One line of work focuses on SLMs that jointly model the distribution of sentences and their structures \cite{chelba-1997-structured, roark-2001-probabilistic, henderson-2004-discriminative, choe-charniak-2016-parsing, kim-etal-2019-unsupervised, dyer-etal-2016-recurrent}. More recent SLMs are mostly based on Transformers. Among them, TGs \cite{sartran-etal-2022-Transformer},  CAGs \cite{yoshida-oseki-2022-composition}, and GPSTs \cite{hu-etal-2024-generative} are closely related to our work as they are constituency-based SLMs with explicit composition. There are also recent studies not covered by our framework: \citet{zhao-etal-2024-dependency} propose dependency-based SLMs, and \citet{qian-etal-2021-structural} and \citet{murty-etal-2023-pushdown} study constituency-based SLMs without explicit composition.
Another line of work augments language models with learnable structures, such as stack-structured memory where syntax patterns are learned from data rather than predefined \cite{joulin2015inferring, yogatama2018memory, dusell2021learning, dusell2023stack}, and learning structural attention patterns \cite{kim2017structured, wang-etal-2019-tree, shen-etal-2021-structformer, shen-etal-2022-unsupervised}. 

\section{Conclusion}
We propose a unified framework for compositional syntactic language models (SLMs) that encompasses four key aspects of design choices. Instances of this framework include not only existing models but also over ten novel variants. Our experiments demonstrate that compositional SLMs outperform the Transformer language model baseline in syntactic generalization and summarization, underscoring the potential of syntactic biases with explicit composition. Furthermore, we comprehensively compare the performance and efficiency of all the SLM variants, resulting in recommendations on the design of compositional SLMs.

\section*{Limitations}
Our framework is currently limited to unlabeled constituency trees and is tested on a relatively small corpus using a GPT-2 backbone due to limited computational resources. Future research will explore other syntactic structures, larger corpora, and more advanced Transformer backbones.
The composition functions employed in our framework show suboptimal performance when modeling 
non-binary trees. This limitation arises from the simplicity of their architecture and the absence of an explicit learning target to guide the composition process beyond the language modeling loss. We identify these as two key areas for enhancing the composition function.

For training and inference, most compositional SLMs are unable to readily leverage recent advancements in Transformer efficiency, such as Flash-Attention \cite{Dao2022FlashAttentionFA}, due to their specific attention patterns. Additionally, we approximate the probability of a sentence by greedily selecting a single-path prefix and marginalizing over 300 sampled trees. Although this approach is commonly used in SLM studies, it is time-consuming and only provides an upper bound for the perplexity metric. We plan to explore more efficient approximation methods in future work.

\section*{Acknowledgement}
This work was supported by the robotic AI-Scientist platform of Chinese Academy of Sciences and by the HPC platform of ShanghaiTech University.

\bibliography{custom,anthology}

\appendix
\section{Binary Trees and Top-down Linearization}\label{app:bi-dn}

When combining top-down linearization and binary trees (i.e., \textbf{\bi-\dn-\#-\#} as defined in section \ref{sec:variants}), for each constituent that is predicted with an opening-nonterminal action, there is no need to predict its closing-nonterminal action, because whenever two of its sub-constituents are generated, the constituent automatically ends. Thus, we may omit closing-nonterminal actions in \textbf{\bi-\dn-\#-\#}. However, this omission would cause (different) problems in both \textbf{\bi-\dn-\ext-\#} and \textbf{\bi-\dn-\internal-\#}. We use input sequence "<bos> ( A ( B C ) )" (whose output sequence is "( A ( B C ) ) <eos>") as our running example to show the problems of such omission:
\begin{itemize}
    \item For \textbf{\bi-\dn-\ext-\#}, if we omit any closing-nonterminal action, the target sequence becomes "( A ( B C <eos>". 
    What is the corresponding input sequence? A natural choice is "<bos> ( A ( B C", but then the two representations composed from (B C) and (A (B C)) never get a chance to be input into the transformer as required in compositional SLMs. If we change the last input from C to one of the composed representations, then C is never input into the transformer, which is intuitively as bad if not worse.
    \item For \textbf{\bi-\dn-\internal-\#}, the input sequence is "<bos> ( A ( B C ) )' ) )'" and the target sequence is "( A ( B C ) \_ ) \_ <eos>". If we omit closing-nonterminal actions, then the input and target sequences become "<bos> ( A ( B C )' )'" and "( A ( B C \_ \_ <eos>". We need full context for <eos> prediction, so the two internal compositions (with attention limited to sub-constituents) of (B C) and (A (B C)) must taken place on the input C and )'. In other words, we have to compose B and C when the input is C. This is problematic because C is not even encoded by the transformer before its composition.
\end{itemize}
Solving all these problems requires non-trivial re-design of SLMs, which may be potentially inconsistent with the framework proposed in the paper. 
We leave this for future work.

\section{Training Details and Variances}
\label{app:variance}

SLMs model linearized trees, which consist of more action tokens than traditional token sequences. To ensure a fair comparison, we train all SLMs with a cutoff length of 2048 and GPT-2 with a cutoff length of 1024. All models are trained with a fixed learning rate of 5e-5
 , and we modify the batch size for each model to fit within the available GPU memory. We spent 4 NVIDIA A6000 GPUs for each training, which lasted approximately 35 hours on average. To address training variance, we provide the evaluation results for \textbf{\bi-\up-\ext-\nm} as an example, as shown in Table \ref{table:variance}, which was trained three times with different random seeds. The variance was found to be small and does not affect the experimental results presented in the paper.

\section{Hyperparameters Selection} \label{app:hyperparameters}
The beam size of 300 is commonly used in previous studies~\cite{qian-etal-2021-structural, murty-etal-2023-pushdown, hu-etal-2024-generative}, so we also fix the beam size at 300. 

It is reasonable to set the maximum number of nonterminals $n_c$ to be the length of a sentence because, for a binary tree, there are exactly $n_{token} - 1$ nonterminals with a sentence of length $n_{token}$. For a non-binary tree, there are even fewer nonterminals. Therefore, the length of a sentence is a good upper bound for the number of nonterminals $n_c$.

We tune the maximum number of consecutive opening-nonterminal actions $p_c$ on the training set of BLLIP{\small-LG}. As the sentences in SG test suites are short in length (usually no more than 20 tokens), so we randomly choose 10 sentences from BLLIP{\small-LG} of no more than 15 tokens. We run word-synchronous beam search to get top-$300$ $p(\mathbf{x,y})$ and approximate $p(\mathbf{x})$ by $\sum_{\mathbf{y}}p(\mathbf{x,y})$. We tune $p_c$ from 2 to 10 and find that when $p_c$ is set to 3, $p(\mathbf{x})$ remains large
for the 10 sentences. So we fix $p_c$ to be 3 for SG evaluation. We also hypothesize that if the sentence is longer, there is likely to be more consecutive opening-nonterminal actions. Therefore, we set $p_c$ to 5 for summarization as the summaries are longer.

\begin{table}
\centering
\resizebox{\columnwidth}{!}{
    \begin{tabular}{l|c|c|c}
        \toprule
         \thead{Model} & \thead{PPL (std)} & \thead{SG (std)} & \thead{R-AVG (std)}\\
         \midrule
         \bi-\up-\ext-\nm & 20.51 (0.09) & 80.1 (0.3) & 20.33 (0.06) \\
         \bottomrule
    \end{tabular}
}
\caption{Mean and Variance.}
\label{table:variance}
\end{table}

\section{More Baselines}\label{app:analysis}
\begin{table}[tb]
  \centering
  \small
  \begin{tabular}{lccc} \toprule
    \thead{Model}           & \thead{PPL$\_\text{single}$$^\dagger$ ($\downarrow$)} & \thead{PPL$^\dagger$ ($\downarrow$)} & \thead{SG ($\uparrow$)} \\\midrule
    \text{GPT2-tree} & \textbf{20.99} & \textbf{19.97} & 73.1 \\
    \text{Left-branching} & 22.20 & 
    21.93 & 54.2 \\
    \text{Right-branching} & 21.34 & 21.22 & 42.5 \\
    \text{GPT2-tree-\bi-\dn} & - & 22.68 & 78.1\\
    \text{GPT2-tree-\bi-\up} & - & 21.13 & \textbf{78.6}\\
    \text{GPT2-tree-\nb-\up} & - & 22.23 & 36.0 \\
 \bottomrule
  \end{tabular}
  \caption{
   Perplexity (PPL) and syntactic generalization (SG) results of GPT2-tree, two trivial tree baselines, and three GPT2-tree variants. PPL\_single means to estimate perplexity with a single linearized tree. $^\dagger$: All the reported PPLs are upper bounds of the true values.
  }
  \label{table:analysis}
\end{table}

In this section, we experiment with a few variants of the GPT2-tree baseline.
\subsection{Trivial Trees}
We first introduce two trivial tree baselines to verify if the better performance of SLMs is achieved by incorporating real syntax or simply by leveraging more computation brought by longer sequences: (i) \textbf{Left-branching}: a SLM that models a linearized binary left-branching tree (e.g., "( ( ( Write an ) essay ) quickly )"), and (ii) \textbf{Right-branching}: a SLM that models a linearized binary right-branching tree (e.g., "( Write ( an  ( essay quickly ) ) )"). 

The performance of the original GPT2-tree and these two new variants on language modeling and syntactic generalization is reported in Table ~\ref{table:analysis}. 
We additionally report \textbf{PPL\_single} that is evaluated with a single proposal tree, which is the gold parse tree for the original GPT2-tree and the left/right-branching tree for the two variants.
As shown in the table, these trivial tree baselines show worse perplexity and significantly worse SG scores. This suggests the importance of incorporating real syntax in the success of SLMs, which cannot be achieved with trivial tree structures.

\subsection{Other Binarization and Linearization Options}

In our main experiments, GPT2-tree is based on top-down linearized non-binary trees. There are three obvious variants of GPT2-tree: (i) GPT2-tree-\textbf{\bi}-\textbf{\dn}: modeling top-down linearized binary trees, (ii) GPT2-tree-\textbf{\bi}-\textbf{\up}: modeling bottom-up linearized binary trees, and (iii) GPT2-tree-\textbf{\nb}-\textbf{\up}: modeling bottom-up linearized non-binary trees. 
As shown in Table~\ref{table:analysis}, GPT2-tree achieves the lowest perplexity and a medium SG score among the four. GPT2-tree-\textbf{\bi}-\textbf{\dn} and GPT2-tree-\textbf{\bi}-\textbf{\up} show higher perplexity, but they gain significant improvement in syntactic generalization, which is consistent with the performance of the corresponding compositional SLMs in section~\ref{exp:sg} (i.e., \textbf{\bi-\#-\#-\#} all achieve impressive SG performance and \textbf{\bi-\up-\ext-\nm} achieves the highest SG score). Furthermore, though outperforming GPT2-tree on SG, these two variants still underperform most of their compositional SLM counterparts (\textbf{\bi-\dn-\#-\#} and \textbf{\bi-\up-\#-\#} respectively), which is consistent with our conclusion that explicit composition is helpful in syntactic generalization. The extremely bad SG performance of GPT2-tree-\textbf{\nb}-\textbf{\up} also coincides with the bad performance of \textbf{\nb-\up-\ext-\nm}, both of which apply additional modules to predict the start of constituents while failing in capturing the complicated interactions among varying numbers of sub-constituents (as also mentioned in section~\ref{exp:sg}).

\section{Syntactic Generalization Details}\label{app:sg}
\begin{figure*}[tb]
    \centering
    \includegraphics[width=\textwidth]{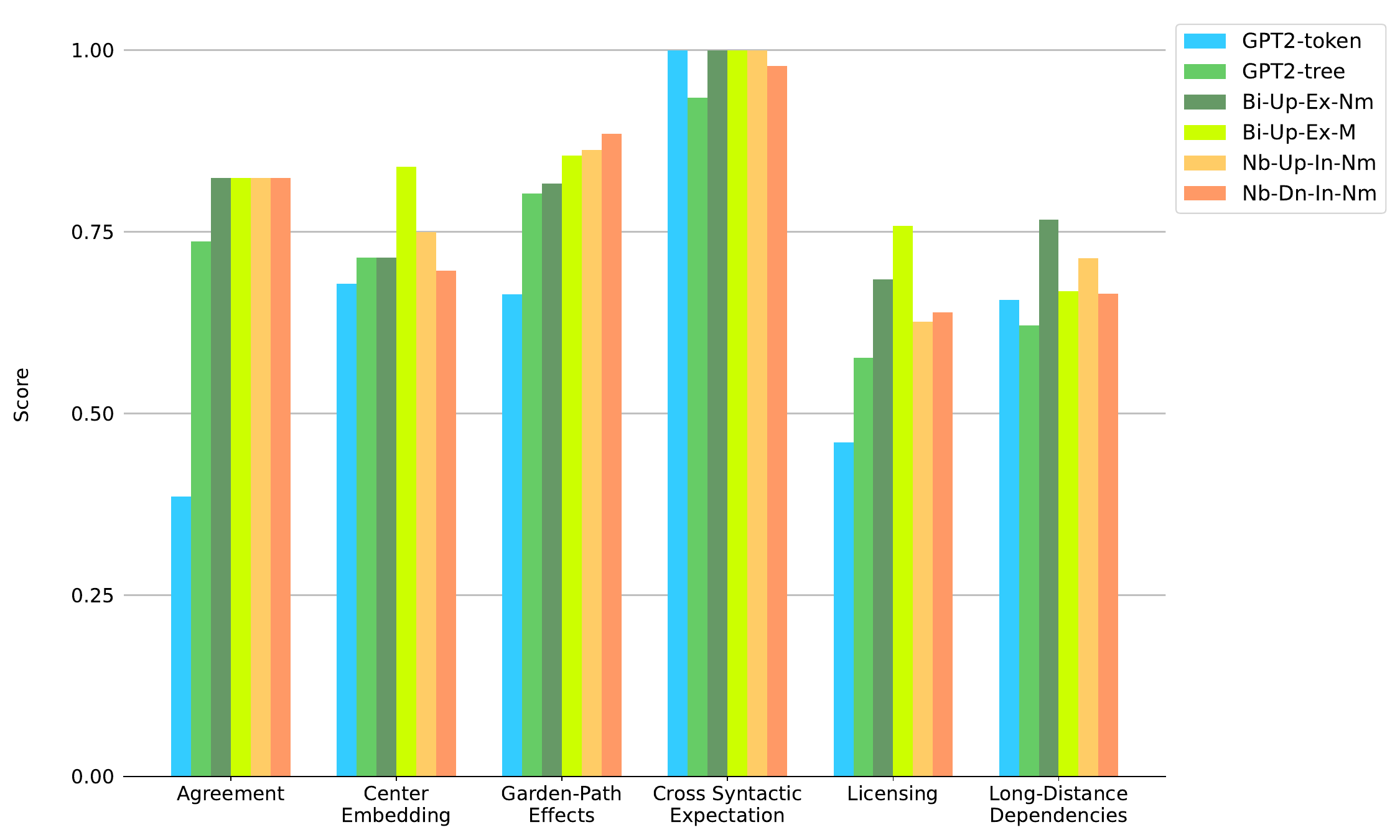}
    \caption{SG scores on each syntactic phenomenon.}
    \label{fig:sg}
\end{figure*}

We present the six detailed syntactic phenomenon scores of two baselines and the four best-performing SLMs in Figure~\ref{fig:sg}. The results show that all four compositional SLMs gain consistent and significant improvement over two baselines on all six circuits. The results again strengthen the conclusion that properly modeling syntax and explicit composition are of help in syntactic generalization.
\section{Number of Model Forward Calls}
\label{app:call}

We record the number of the main Transformer forward calls of running word-synchronous beam search on a sentence of twenty tokens and present it in Table~\ref{table:calls} as a supplement for inference time results. We exclude the forward calls of the external composition function, as it has a much smaller parameter size and each forward pass takes significantly less time. 
\begin{table}
  \centering
  \small
  \begin{tabular}{lcccc}
    \toprule
    \textbf{Model}           & \textbf{bsz-10} & \textbf{bsz-30} & \textbf{bsz-100} & \textbf{bsz-300} \\
    \midrule
    \text{GPT2-tree} & \cellcolor[rgb]{0.13725490196078433, 0.8274509803921569, 0.13725490196078433}{147} & \cellcolor[rgb]{0.49019607843137253, 0.8980392156862745, 0.49019607843137253}{202} & \cellcolor[rgb]{0.5882352941176471, 0.9176470588235295, 0.5882352941176471}{236} & \cellcolor[rgb]{0.5686274509803921, 0.9137254901960784, 0.5686274509803921}{250} \\
    {\bi-\up-\ext-\nm} & \cellcolor[rgb]{0.29411764705882354, 0.8588235294117648, 0.29411764705882354}{165} &  \cellcolor[rgb]{0.2549019607843137, 0.8509803921568628, 0.2549019607843137}{175} &   \cellcolor[rgb]{0.0784313725490196, 0.8156862745098039, 0.0784313725490196}{179} &   \cellcolor[rgb]{0.0, 0.8, 0.0}{182} \\
    {\bi-\up-\ext-\m} &  \cellcolor[rgb]{0.29411764705882354, 0.8588235294117648, 0.29411764705882354}{165} &  \cellcolor[rgb]{0.21568627450980393, 0.8431372549019608, 0.21568627450980393}{170} &  \cellcolor[rgb]{0.09803921568627451, 0.819607843137255, 0.09803921568627451}{180} &  \cellcolor[rgb]{0.0, 0.8, 0.0}{183} \\
    {\bi-\up-\internal-\nm} &  \cellcolor[rgb]{1.0, 0.8941176470588235, 0.8941176470588235}{309} &   \cellcolor[rgb]{1.0, 0.8823529411764706, 0.8823529411764706}{329} &  \cellcolor[rgb]{1.0, 0.9019607843137255, 0.9019607843137255}{341} &   \cellcolor[rgb]{1.0, 0.9215686274509804, 0.9215686274509804}{345} \\
    {\bi-\up-\internal-\m} &  \cellcolor[rgb]{1.0, 0.8941176470588235, 0.8941176470588235}{309} &   \cellcolor[rgb]{1.0, 0.8862745098039215, 0.8862745098039215}{327} & \cellcolor[rgb]{1.0, 0.9058823529411765, 0.9058823529411765}{339} &   \cellcolor[rgb]{1.0, 0.9098039215686275, 0.9098039215686275}{351} \\
    {\bi-\dn-\ext-\nm} &  \cellcolor[rgb]{0.3137254901960784, 0.8627450980392157, 0.3137254901960784}{167} &   \cellcolor[rgb]{0.27450980392156865, 0.8549019607843138, 0.27450980392156865}{178} &  \cellcolor[rgb]{0.45098039215686275, 0.8901960784313726, 0.45098039215686275}{221} &  \cellcolor[rgb]{0.49019607843137253, 0.8980392156862745, 0.49019607843137253}{239} \\
    {\bi-\dn-\ext-\m} &  \cellcolor[rgb]{0.23529411764705882, 0.8470588235294118, 0.23529411764705882}{158} &  \cellcolor[rgb]{0.47058823529411764, 0.8941176470588236, 0.47058823529411764}{201} &  \cellcolor[rgb]{0.37254901960784315, 0.8745098039215686, 0.37254901960784315}{212} &  \cellcolor[rgb]{0.47058823529411764, 0.8941176470588236, 0.47058823529411764}{237} \\
    {\bi-\dn-\internal-\nm} &  \cellcolor[rgb]{1.0, 0.9647058823529412, 0.9647058823529412}{267} &  \cellcolor[rgb]{1.0, 0.9333333333333333, 0.9333333333333333}{300} &  \cellcolor[rgb]{1.0, 0.9058823529411765, 0.9058823529411765}{339} &   \cellcolor[rgb]{1.0, 0.8117647058823529, 0.8117647058823529}{407} \\
    {\bi-\dn-\internal-\m} &  \cellcolor[rgb]{1.0, 0.9490196078431372, 0.9490196078431372}{275} &  \cellcolor[rgb]{1.0, 0.9411764705882354, 0.9411764705882354}{295} &   \cellcolor[rgb]{1.0, 0.9333333333333333, 0.9333333333333333}{324} &   \cellcolor[rgb]{1.0, 0.8196078431372549, 0.8196078431372549}{403} \\
    {\nb-\up-\ext-\nm} & \cellcolor[rgb]{0.7254901960784312, 0.9450980392156862, 0.7254901960784312}{215} & \cellcolor[rgb]{1.0, 0.9882352941176471, 0.9882352941176471}{268} &  \cellcolor[rgb]{1.0, 0.9254901960784314, 0.9254901960784314}{328} &   \cellcolor[rgb]{1.0, 0.9019607843137255, 0.9019607843137255}{356} \\
    {\nb-\up-\ext-\m} & \cellcolor[rgb]{1.0, 0.7607843137254902, 0.7607843137254902}{385} &  \cellcolor[rgb]{1.0, 0.7764705882352941, 0.7764705882352941}{390} &  \cellcolor[rgb]{1.0, 0.8, 0.8}{399} &  \cellcolor[rgb]{1.0, 0.8156862745098039, 0.8156862745098039}{405} \\
    {\nb-\up-\internal-\nm} &  \cellcolor[rgb]{1.0, 0.48235294117647054, 0.48235294117647054}{547} &  \cellcolor[rgb]{1.0, 0.4352941176470587, 0.4352941176470587}{587} &  \cellcolor[rgb]{1.0, 0.4235294117647058, 0.4235294117647058}{617} &   \cellcolor[rgb]{1.0, 0.3764705882352941, 0.3764705882352941}{659} \\
    {\nb-\up-\internal-\m} &  \cellcolor[rgb]{1.0, 0.2, 0.2}{713} & \cellcolor[rgb]{1.0, 0.2, 0.2}{725} & \cellcolor[rgb]{1.0, 0.2, 0.2}{747} &   \cellcolor[rgb]{1.0, 0.2, 0.2}{763} \\
    {\nb-\dn-\ext-\nm} &  \cellcolor[rgb]{0, 0.8, 0}{129} &  \cellcolor[rgb]{0.0, 0.8, 0.0}{145} &   \cellcolor[rgb]{0.0, 0.8, 0.0}{168} &   \cellcolor[rgb]{0.058823529411764705, 0.8117647058823529, 0.058823529411764705}{190} \\
    {\nb-\dn-\ext-\m} & \cellcolor[rgb]{0.2549019607843137, 0.8509803921568628, 0.2549019607843137}{159} &  \cellcolor[rgb]{0.3137254901960784, 0.8627450980392157, 0.3137254901960784}{182} & \cellcolor[rgb]{0.43137254901960786, 0.8862745098039216, 0.43137254901960786}{219} &   \cellcolor[rgb]{0.5686274509803921, 0.9137254901960784, 0.5686274509803921}{248} \\
    {\nb-\dn-\internal-\nm} &  \cellcolor[rgb]{1.0, 0.9411764705882354, 0.9411764705882354}{280} &   \cellcolor[rgb]{1.0, 0.8666666666666667, 0.8666666666666667}{339} &   \cellcolor[rgb]{1.0, 0.7960784313725491, 0.7960784313725491}{402} &   \cellcolor[rgb]{1.0, 0.7803921568627451, 0.7803921568627451}{425} \\
    {\nb-\dn-\internal-\m} &  \cellcolor[rgb]{1.0, 0.8901960784313725, 0.8901960784313725}{310} &  \cellcolor[rgb]{1.0, 0.8431372549019608, 0.8431372549019608}{352} & \cellcolor[rgb]{1.0, 0.7725490196078432, 0.7725490196078432}{416} & \cellcolor[rgb]{1.0, 0.7647058823529411, 0.7647058823529411}{435} \\
    \bottomrule
  \end{tabular}
  \caption{
   The number of model forward calls.
  }
  \label{table:calls}
\end{table}
\section{Discussion on Inference Time of Modeling Linearized Non-binary Trees}
\label{app:inference}
As shown in Table \ref{table:calls}, compared to \textbf{\bi-\#-\#-\#}, \textbf{\nb-\#-\#-\#} generally incurs more forward calls (except in the case of \textbf{\bi-\dn-\ext-\nm} vs. \textbf{\nb-\dn-\ext-\nm}). This discrepancy arises from word-synchronous beam search, where at each synchronous step, the top-$k$ beams of non-binary trees tend to exhibit wide variation in the number of compositions, which is significantly higher than in binary trees. Consequently, during each pair of synchronous steps, some beams have few composed constituents, while others have many more. Those with fewer composed constituents perform multiple compositions, while others immediately generate a new token and wait for synchronization, leading to a consistently high number of forward calls during each pair of the synchronous steps, a phenomenon absent in binary settings. This phenomenon is more pronounced in \textbf{\nb-\up-\#-\#} than in \textbf{\nb-\dn-\#-\#}, resulting in more forward calls for \textbf{\nb-\up-\#-\#}. We consider this an intrinsic challenge of applying word-synchronous beam search to SLMs that model non-binary trees.

\section{Generation Efficiency}\label{app:eff}
\begin{table}[ht]
  \centering
  \small
  \begin{tabular}{lc}
    \toprule
    \textbf{Model}  & \thead{Time\_Per\_Token (ms)} \\
    \midrule
    \text{GPT2-token} & 5.6 \\
    \text{GPT2-tree} & 50.9 \\
    {\bi-\up-\ext-\nm} & 26.1\\
    {\bi-\up-\ext-\m} &  27.8\\
    {\bi-\up-\internal-\nm} & 57.1\\
    {\bi-\up-\internal-\m} & 64.6\\
    {\nb-\up-\internal-\nm} & 65.5\\
    {\nb-\up-\internal-\m} &  71.1\\
    {\nb-\dn-\internal-\nm} & 70.2 \\
    {\nb-\dn-\internal-\m} &  84.1 \\
    \bottomrule
  \end{tabular}
  \caption{
   The generation time (lower is better).
  }
  \label{table:gen_eff}
\end{table}
Experiments presented in this section are conducted with a single A800 GPU.

For a fair comparison between GPT2-token and SLMs, we use exactly the same setup as in the summarization experiment (section~\ref{exp:xsum}), applying word-synchronous beam search to top-$k$ random sampling (using a beam size of 2 and $k=2$) for SLMs and use top-$k$ random sampling ($k=2$) for GPT2-token. We randomly pick 100 prompts with an average length of 106.16 (tokens) and run all the models to generate 100 tokens with each prompt. We report the average generation time per token in Table~\ref{table:gen_eff}. The results show that all the SLMs show significantly lower efficiency than GPT2-token due to additionally encoding and generating syntactic structures, which is an intrinsic limitation of SLMs. On the other hand, $\ext$ shows higher efficiency compared with $\internal$, which is consistent with the results in section~\ref{exp:eff}. 
\end{document}